%% file: camera_ready_preprint.tex
\documentclass[10pt,twocolumn,letterpaper]{article}

\usepackage[algorithms]{wacv}      % To produce the REVIEW version for the algorithms track
\usepackage{multicol}
\usepackage{amsmath,amssymb,amsfonts}
\usepackage{algorithmic}
\usepackage{graphicx}
\usepackage{textcomp}
\usepackage{booktabs,multirow,adjustbox}
\usepackage{array,makecell,graphicx,subcaption,booktabs}

\usepackage{booktabs}
\usepackage{multirow}
\usepackage{adjustbox}
\usepackage{colortbl}
\usepackage{xcolor}
\definecolor{psnrbg}{RGB}{245,245,245}
\definecolor{vifbg}{RGB}{235,245,255}
\definecolor{fsimbg}{RGB}{245,235,255}
\input{preamble}

\definecolor{wacvblue}{rgb}{0.21,0.49,0.74}
\usepackage[pagebackref,breaklinks,colorlinks,allcolors=wacvblue]{hyperref}

\def\wacvPaperID{1107} % *** Enter the WACV Paper ID here
\def\confName{WACV}
\def\confYear{2027}

\title{MRI Reconstruction Risk Estimation from Inference Dynamics}

\author{Tamir Shor\\
Technion\\
{\tt\small tamir.shor@campus.technion.ac.il}
\and
Alex Bronstein\\
Institute of Science and Technology Austria\\
{\tt\small alexander.bronstein@ist.ac.at}
}
\begin{document}
\maketitle

\begin{abstract}
Iterative MRI reconstructors produce intermediate estimates before returning a final image. We study these inference trajectories as a model-internal representation of realized reconstruction risk, measured by log-NMSE. Model-native states are projected to magnitude images, summarized by a fixed 16-dimensional descriptor, and mapped to risk using a source-trained ridge readout. Across three reconstruction backbones, trajectory features are strongly predictive of reconstruction risk on held-out source data. Under zero-shot target shift, their behavior is backbone-dependent, exhibiting both preserved positive association and readout reversal. Conditional analyses show that trajectory information can remain associated with error beyond an initialization-only predictor. A prefix analysis further shows that useful trajectory information can emerge before reconstruction completes. Repeated-mask and low-shot calibration studies indicate that trajectories primarily encode persistent case difficulty, while their orientation and calibration can change under domain shift. Code is available at \href{https://github.com/tamirshor7/MRI-Risk-From-Inference-Dynamics}{\textcolor{purple}{https://github.com/tamirshor7/MRI-Risk-From-Inference-Dynamics}}
\end{abstract}

\section{Introduction}
\label{sec:introduction}

Accelerated magnetic resonance imaging reduces scan time by measuring part of $k$-space and reconstructing the missing information computationally. Parallel imaging and compressed sensing provide established formulations for this inverse problem \cite{pruessmann1999sense,griswold2002grappa,lustig2007sparse}. Learned reconstruction methods combine trainable priors with the MRI forward model and data consistency \cite{hammernik2018variational,aggarwal2019modl,sriram2020varnet,ramzi2020xpdnet,xin2023promptmr,xin2024rethinking}. These methods can return visually plausible images with structured residual error. Deployment therefore requires a case-level estimate of reconstruction error when the fully sampled reference is unavailable.

Fully sampled references are generally unavailable at inference. Existing risk estimators use the final reconstruction, stochastic predictions, learned error maps, ensembles, or calibrated interval widths \cite{hu2021predict,edupuganti2021uncertainty,kustner2024predictive,khawaled2024npbrec,romano2019cqr,angelopoulos2022im2im}. Strong final-image predictors can estimate reconstruction error. Iterative reconstructors also expose a sequence of internal estimates that records how the model reached its output. This sequence provides a separate source of evidence about update magnitude, convergence shape, cancellation, and motion near the endpoint. Trajectory statistics can also become available before reconstruction finishes, enabling early triage.

A controlled-corruption study first showed that intermediate-state geometry changes reproducibly with degradation severity while failing a causal correction probe. This separation motivates an observational risk estimator. Full directional and intervention analyses are provided in Supplementary Sec. S10.

We project the native states of an iterative MRI reconstructor to a common image space and summarize the resulting path with 16 pre-specified geometric features. A low-capacity ridge model maps these features to log-NMSE. The main generality experiment applies exactly the same descriptor and evaluation protocol across three substantially different iterative reconstruction backbones on matched fastMRI-brain source data and M4Raw target data. It compares trajectory risk with a Hu-style final-image error predictor, a scalar final-image predictor, conformalized uncertainty scores, an initialization-only control, and inexpensive physics-side diagnostics. The target evaluation is zero-shot. Risk models are fitted on source data, and M4Raw labels are used only for evaluation.

The results separate representation informativeness from readout transfer. All three trajectory predictors are strong on held-out source data. Under zero-shot target shift, one readout retains strong positive association, one remains positive but attenuated, and one reverses during severe reconstruction shift. Initialization and endpoint controls show that trajectory information can remain conditionally associated with error. A prefix experiment supplies most target ranking information by iteration 12 of 25. A repeated-mask experiment further shows that the trajectory mainly measures persistent case difficulty across physical volumes. Affine target readouts recover useful prediction from reversed coordinates in two stress tests.

The paper makes three contributions. First, it defines a compact image-space trajectory representation shared across three substantially different iterative reconstruction backbones. Second, it evaluates this representation against initialization, learned final-image, uncertainty, and physics-side controls under a matched source and target protocol. Third, it shows that trajectory risk mainly reflects persistent case difficulty, can become informative before inference completes, and can retain target-risk structure after domain shift changes its source-trained orientation.

\begin{figure*}[t]
    \centering
    \includegraphics[width=0.92\textwidth]{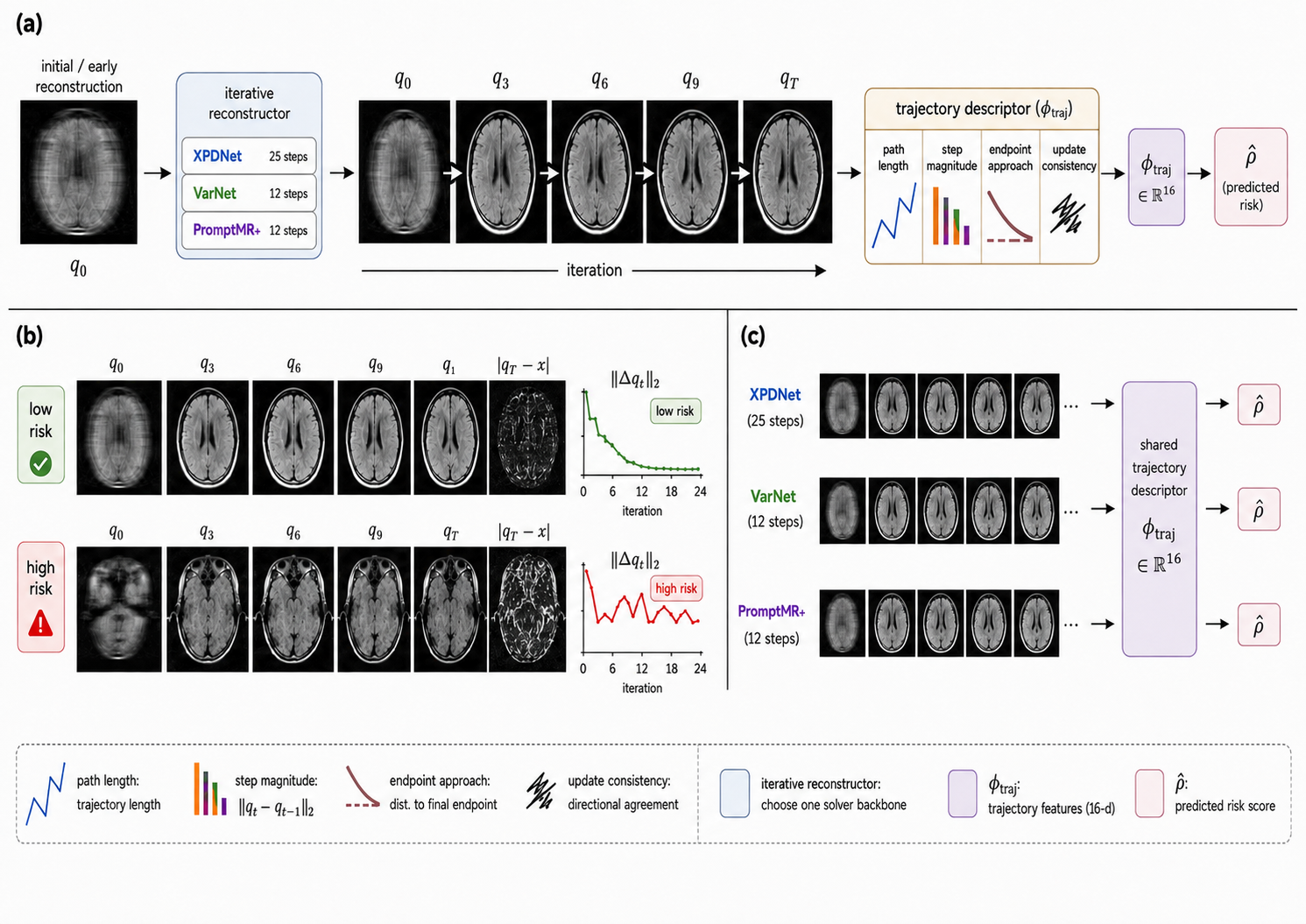}
    \caption{Method overview - (a) Intermediate reconstructions from an iterative solver are projected to a common image-space trajectory, summarized by a 16-dimensional descriptor, and mapped to predicted reconstruction risk. (b) Low- and high-risk cases can have plausible final reconstructions but exhibit different trajectory dynamics and residual error. (c) The same trajectory descriptor is applied across reconstruction backbones.}
    \label{fig:overview}
\end{figure*}

\section{Related Work}
\label{sec:related_work}

\textbf{Learned iterative reconstruction.}
Unrolled networks translate iterations of an optimization procedure into trainable stages. Variational networks learn regularization within a variational update \cite{hammernik2018variational}. MoDL alternates learned regularization with model-based data consistency \cite{aggarwal2019modl}. E2E-VarNet integrates sensitivity estimation and multi-coil reconstruction \cite{sriram2020varnet}. XPDNet is a primal-dual-inspired cross-domain unrolled reconstruction network \cite{ramzi2020xpdnet}. PromptMR exploits adjacent context through prompted reconstruction modules \cite{xin2023promptmr}, while PromptMR+ enhances gradient and information flow within and across unrolled stages and improves adjacent-information-based sensitivity estimation \cite{xin2024rethinking}. These architectures expose intermediate estimates during their standard forward pass. We study the geometry of those estimates as a risk modality.

\textbf{Uncertainty and error prediction.}
Monte Carlo dropout and deep ensembles approximate predictive uncertainty through repeated inference \cite{gal2016dropout,lakshminarayanan2017deep}. MRI methods include variational uncertainty, deep ensembles under acquisition shifts, and parameter-posterior sampling \cite{edupuganti2021uncertainty,kustner2024predictive,khawaled2024npbrec}. SURE and generalized SURE estimate expected risk through assumptions on the measurement process \cite{stein1981sure,eldar2009gsure}. Hu et al. propose a two-step framework that separately trains a reconstruction model and an error-prediction model \cite{hu2021predict}. DEUP studies secondary error prediction from out-of-sample errors \cite{lahlou2023deup}. Our target is the realized scalar error of one deterministic reconstruction. The contribution lies in using solver dynamics as the predictor input and in testing whether they add information beyond strong endpoint models.

\textbf{Calibration and distribution shift.}
Conformalized quantile regression constructs marginal prediction intervals from a labeled calibration set \cite{romano2019cqr}. Risk-controlling methods extend calibration to image-valued outputs and general losses \cite{angelopoulos2022im2im,bates2021risk}. Imaging applications include conformal bounds for full-reference quality metrics \cite{wen2025friq}. Giannakopoulos et al. construct pixel-wise uncertainty intervals for accelerated MRI reconstruction \cite{giannakopoulos2026pixelwise}, while our objective is one scalar realized case-level risk coordinate from deterministic solver dynamics. Reconstruction robustness studies report sensitivity to anatomy, contrast, acceleration, masks, and noise \cite{darestani2022testtime,fujita2024robustness}. We hold each reconstructor fixed and distinguish the trajectory representation from the scalar readout that interprets it. This distinction is evaluated directly through signed zero-shot transfer and low-shot target calibration.

\section{Method}
\label{sec:method}

\subsection{Image-Space Reconstruction Trajectories}
\label{sec:trajectory}

Let $x\in\mathbb{C}^{H\times W}$ be a fully sampled image, $S_c$ the sensitivity of coil $c$, $M$ a Cartesian mask, and $\mathcal{F}$ the centered Fourier transform. The measured multi-coil data are
\begin{equation}
    y=\mathcal{A}_{M}(x)+\xi,
    \quad
    \mathcal{A}_{M}(x)=\{M\odot\mathcal{F}(S_c\odot x)\}_{c=1}^{C},
    \label{eq:forward}
\end{equation}
where $\xi$ is measurement noise. A frozen iterative reconstructor maintains a native state $z_t$ and applies
\begin{equation}
    z_{t+1}=G_{\theta,t}(z_t,y,M),
    \quad t=0,\ldots,T-1.
    \label{eq:update}
\end{equation}

The state may contain images, multi-coil $k$-space, sensitivity-weighted features, or primal and dual variables. A model-consistent projection $\Pi_t$ maps each state to the magnitude image implied by the solver,
\begin{equation}
    q_t=\Pi_t(z_t)\in\mathbb{R}_{+}^{H_r\times W_r},
    \quad
    \mathcal{T}(y)=(q_0,\ldots,q_T).
    \label{eq:trajectory}
\end{equation}
The projection uses the reconstruction operations native to each model, including inverse Fourier transformation, sensitivity reduction, and coil combination where required. It gives the model initialization at $q_0$ and the returned magnitude image at $q_T$.

The trajectory exposes update scale, cancellation, and relations among updates. We summarize these observable quantities without using the unavailable reference image.

\subsection{Descriptor and Risk Readout}
\label{sec:descriptor}

For an image $u$, define $\rms(u)=\|u\|_2/\sqrt{H_rW_r}$ and $s_0=\rms(q_0)+\epsilon$. Normalized step magnitudes are
\begin{equation}
    d_t=\frac{\rms(q_t-q_{t-1})}{s_0},
    \quad t=1,\ldots,T.
    \label{eq:steps}
\end{equation}
The descriptor $\phi_{\mathrm{traj}}\in\mathbb{R}^{16}$ contains path length, endpoint displacement, path efficiency, five step-size summaries, two log-step slopes, three endpoint-approach summaries, and three consecutive-update cosine summaries. Early and late windows use the first and last thirds. Regression uses raw indices $1,\ldots,T$. Endpoint summaries use arithmetic means, and a degenerate update pair receives cosine zero. Complete definitions are given in Supplementary Sec. S1.

Three representative quantities are
\begin{equation}
    L=\sum_{t=1}^{T}d_t,
    \quad
    D=\frac{\rms(q_T-q_0)}{s_0},
    \quad
    \eta=\frac{D}{L+\epsilon},
    \label{eq:path_geometry}
\end{equation}
which measure total motion, net displacement, and path directness. Endpoint approach is described by
\begin{equation}
    a_t=\frac{\rms(q_T-q_t)}{\rms(q_T)+\epsilon},
    \quad t=0,\ldots,T,
    \label{eq:endpoint_distance}
\end{equation}
and consecutive-update agreement is
\begin{equation}
    c_t=\frac{\langle\Delta_t,\Delta_{t-1}\rangle}
    {\|\Delta_t\|_2\|\Delta_{t-1}\|_2},
    \quad t=2,\ldots,T.
    \label{eq:update_cosine}
\end{equation}
The descriptor combines levels, temporal slopes, early-to-late ratios, and dispersion. It contains no target image, error value, acquisition label, patient identifier, or target-domain statistic.

The same descriptor is applied to 25 XPDNet steps, 12 VarNet cascades, and 12 PromptMR+ cascades. Each architecture has its own state projection and source standardization. The earlier single-coil eight-cascade reconstruction model used in the discovery and repeated-mask analyses, hereafter the \emph{discovery model}, retains a distinct pre-specified 16-dimensional descriptor that summarizes the same geometric families. The discovery model is excluded from the shared-descriptor backbone-generalization claim.

\begin{table}[t]
    \centering
    \small
    \caption{Matched trajectory construction. Projection is model-specific, while the 16 feature functions are shared.}
    \label{tab:backbones}
    \setlength{\tabcolsep}{3.5pt}
    \begin{tabular}{lcl}
        \toprule
        Backbone & Steps & Projected state \\
        \midrule
        XPDNet & 25 & primal-dual image estimate \\
        VarNet & 12 & sensitivity-reduced cascade output \\
        PromptMR+ & 12 & central image with five-slice context \\
        \bottomrule
    \end{tabular}
\end{table}

For source feature mean $\mu$ and scale $\sigma$, the risk predictor is
\begin{equation}
    \widehat{r}(y)=b+w^{\top}
    \left(\frac{\phi_{\mathrm{traj}}(\mathcal{T}(y))-\mu}{\sigma}\right),
    \quad
    r(y)=\log\nmse(q_T,x).
    \label{eq:risk}
\end{equation}
Ridge regularization is selected using source data. The reconstructor, projection, feature definition, normalization, and coefficients are frozen for zero-shot evaluation. The score uses one deterministic reconstruction pass.

Feature standardization is fitted on the source training partition. Features with source SD at most $10^{-10}$ are removed. The log-NMSE target is centered on the same partition. The ridge objective is
\begin{equation}
    \min_{w,b}\sum_{i\in\mathcal{D}_{\mathrm{src}}}
    \left(r_i-b-w^{\top}\widetilde{\phi}_i\right)^2
    +\lambda\|w\|_2^2,
    \label{eq:ridge}
\end{equation}
where $\widetilde{\phi}_i$ is the standardized descriptor. This readout has 16 coefficients before removal of degenerate source features. Its capacity is small relative to the image-based comparators.

\subsection{Comparators and Evaluation}
\label{sec:comparators}

We compare the trajectory readout with four source-trained estimators. Our Hu-style comparator adapts the two-step error-prediction idea of Hu et al. to a final-reconstruction input \cite{hu2021predict}. Its pooled error-map magnitude is used as a case-level score. ResNet maps the final magnitude image directly to scalar log-NMSE. CQR fits lower and upper image quantiles and uses normalized mean calibrated interval width \cite{angelopoulos2022im2im,romano2019cqr}. RM-RMS uses normalized RMS residual-magnitude interval width following image-to-image risk control \cite{angelopoulos2022im2im}. CQR and RM-RMS are evaluated as risk scores through correlation with log-NMSE. We also evaluate endpoint amplitude and the final sampled-space data-consistency residual as inexpensive controls.

The comparators answer different questions. Hu-style and ResNet test whether a learned endpoint representation is sufficient for error prediction. CQR and RM-RMS test whether uncertainty width orders realized scalar error. The data-consistency residual tests whether a physics-side diagnostic carries deployment information. Their raw values need not share the units or calibration of the trajectory predictor, so the common primary metrics are signed Pearson and Spearman correlation. RMSE is additionally reported for models trained as point predictors.

Pearson correlation $r$ measures signed affine association, Spearman correlation $\rho_s$ measures signed rank association, and RMSE measures calibration for point predictors. A negative target correlation means that the source-trained risk orientation has reversed. Single-method uncertainty for XPDNet, VarNet, and PromptMR+ uses 10,000 patient-cluster bootstrap replicates. Source and target resampling units are patients, and paired comparisons use matched resamples. M4Raw slices are aggregated to scan-level scores before patient resampling. Supplementary Secs. S2--S4 report 95\% percentile intervals and paired comparisons.

Complementarity with the Hu-style predictor is measured through partial correlation with true log-NMSE after linearly residualizing the other score. As an initialization-only control, we train the same scalar ResNet architecture on $q_0$ using identical source-only fitting and zero-shot target evaluation. Conditional analyses residualize trajectory and $q_0$ against one another. We also report a source-only, meta-out-of-fold XPDNet fusion. VarNet and PromptMR+ fusion models use a different stacking protocol, so fusion is kept outside the primary matched tables and reported in Supplementary Sec. S4. The severe-shift calibration experiment retains the trajectory coordinate and fits only
\begin{equation}
    \widehat{r}_{\mathrm{target}}=a+b\widehat{r}_{\mathrm{source}}
    \label{eq:calibration}
\end{equation}
from a small labeled target cohort.

\section{Results}
\label{sec:results}

\subsection{Matched Evaluation Protocol}
\label{sec:protocol}

The matched source dataset is fastMRI brain \cite{muckley2021fastmri}. It contains 3676 evaluated slices, with 2944 source-fit rows from 736 patients and 732 held-out rows from 183 patients. Four central slices are evaluated per volume. The target dataset is M4Raw FLAIR \cite{lyu2023m4raw}, with 400 slices from 100 scans and 25 patients. Four slices are averaged into each scan-level target score before patient-cluster evaluation. XPDNet, VarNet, and PromptMR+ use the same selected volumes, slices, mask semantics, risk target, and 16 feature definitions.

All risk estimators are fitted on fastMRI brain. M4Raw labels are used only for evaluation in the matched zero-shot experiment. The architecture comparison is therefore controlled at the risk-estimator level while preserving each reconstructor's native computation. Supplementary Secs. S2--S3 document exact score definitions and full uncertainty intervals.

\subsection{Source Prediction and Zero-Shot Transfer}
\label{sec:matched_results}

Table~\ref{tab:source_matched} reports held-out source results. On XPDNet, trajectory reaches $r=0.804\pm0.027$ and $\rho_s=0.807\pm0.026$. Hu-style is similar in Pearson correlation at $0.805\pm0.028$ and lower in rank correlation at $0.774\pm0.030$, while ResNet is strongest on the source holdout at $r=0.911\pm0.014$ and $\rho_s=0.900\pm0.015$. Trajectory Pearson correlation is $0.849\pm0.024$ for VarNet and $0.906\pm0.016$ for PromptMR+. The same descriptor is therefore informative for all three backbones. Its relative position among comparators varies. CQR is close to trajectory for VarNet and PromptMR+, and RM-RMS is negatively associated with source error for VarNet and PromptMR+.

\begin{table*}[t]
    \centering
    \footnotesize
    \caption{Matched fastMRI-brain source holdout. Values are signed correlations with log-NMSE and are reported as estimate $\pm$ patient-cluster bootstrap SD from 10,000 replicates for all three backbones. CQR uses normalized mean interval width. RM-RMS uses normalized RMS interval width. $^\dagger$ The table reports the pre-specified source-selected ResNet realization. Additional training seeds show substantial target variability in Supplementary Table S15.}
    \label{tab:source_matched}
    \setlength{\tabcolsep}{4.1pt}
    \begin{tabular}{lcccccc}
        \toprule
        & \multicolumn{2}{c}{XPDNet} & \multicolumn{2}{c}{VarNet} & \multicolumn{2}{c}{PromptMR+} \\
        \cmidrule(lr){2-3}\cmidrule(lr){4-5}\cmidrule(lr){6-7}
        Method & $r$ & $\rho_s$ & $r$ & $\rho_s$ & $r$ & $\rho_s$ \\
        \midrule
        \textbf{Trajectory} & $+0.804\pm0.027$ & $+0.807\pm0.026$ & $+0.849\pm0.024$ & $+0.834\pm0.022$ & $+0.906\pm0.016$ & $+0.895\pm0.017$ \\
        Hu-style & $+0.805\pm0.028$ & $+0.774\pm0.030$ & $+0.757\pm0.029$ & $+0.734\pm0.033$ & $+0.616\pm0.039$ & $+0.563\pm0.045$ \\
        ResNet$^\dagger$ & $+0.911\pm0.014$ & $+0.900\pm0.015$ & $+0.588\pm0.050$ & $+0.551\pm0.051$ & $+0.187\pm0.052$ & $+0.136\pm0.061$ \\
        CQR & $+0.790\pm0.031$ & $+0.786\pm0.028$ & $+0.819\pm0.022$ & $+0.823\pm0.024$ & $+0.855\pm0.018$ & $+0.860\pm0.019$ \\
        RM-RMS & $-0.023\pm0.062$ & $-0.022\pm0.070$ & $-0.246\pm0.065$ & $-0.282\pm0.068$ & $-0.115\pm0.054$ & $-0.182\pm0.065$ \\
        \bottomrule
    \end{tabular}
\end{table*}

Table~\ref{tab:source_rmse} reports source log-NMSE RMSE for the calibrated point predictors. For VarNet, trajectory achieves $0.193\pm0.011$, compared with $0.240\pm0.010$ for Hu-style and $0.324\pm0.014$ for ResNet. For PromptMR+, trajectory achieves $0.169\pm0.010$, compared with $0.316\pm0.017$ and $0.492\pm0.022$. XPDNet exhibits a different ordering. ResNet has the lowest RMSE at $0.188\pm0.007$, Hu-style reaches $0.228\pm0.011$, and trajectory reaches $0.231\pm0.014$. Trajectory is therefore the best-calibrated source point predictor for VarNet and PromptMR+, while ResNet is strongest for XPDNet.

\begin{table}[t]
    \centering
    \small
    \caption{Source log-NMSE RMSE for matched point predictors. Values are estimate $\pm$ patient-cluster bootstrap SD from 10,000 replicates. Lower is better.}
    \label{tab:source_rmse}
    \setlength{\tabcolsep}{3.6pt}
    \begin{tabular}{lccc}
        \toprule
        Backbone & Trajectory & Hu-style & ResNet$^\dagger$ \\
        \midrule
        XPDNet & $0.231\pm0.014$ & $0.228\pm0.011$ & $0.188\pm0.007$ \\
        VarNet & $0.193\pm0.011$ & $0.240\pm0.010$ & $0.324\pm0.014$ \\
        PromptMR+ & $0.169\pm0.010$ & $0.316\pm0.017$ & $0.492\pm0.022$ \\
        \bottomrule
    \end{tabular}
\end{table}

Table~\ref{tab:target_matched} gives signed M4Raw scan-level correlations. PromptMR+ preserves a strong positive target association. Its trajectory reaches $r=0.856\pm0.043$ and $\rho_s=0.852\pm0.052$. The paired Pearson estimate narrowly favors trajectory over Hu-style by $0.209\pm0.118$, with 95\% CI $[0.002,0.457]$, while the Spearman interval includes zero. Paired differences over the reported ResNet realization, CQR, and RM-RMS are $0.564\pm0.220$, $0.434\pm0.161$, and $0.761\pm0.135$ in Pearson correlation, with both metric intervals excluding zero. Additional ResNet seeds show substantial target sensitivity, so the principal conclusion does not rely on that single endpoint realization. The three-seed analysis is reported in Supplementary Table S15.

XPDNet retains a positive zero-shot target association, with $r=0.630\pm0.086$ and $\rho_s=0.662\pm0.090$. Hu-style is statistically competitive at $r=0.651\pm0.099$ and $\rho_s=0.647\pm0.113$. Paired trajectory-minus-Hu-style intervals include zero in both metrics. Trajectory exceeds CQR and RM-RMS in the paired comparisons, with Pearson difference intervals $[0.005,0.817]$ and $[0.078,0.636]$.

VarNet shows a different outcome. The trajectory correlations are $r=-0.935\pm0.020$ and $\rho_s=-0.924\pm0.030$. These values are retained with their negative sign because they represent a failed zero-shot risk orientation. This backbone experiences a substantially more severe reconstruction shift than the other matched models. Median slice NMSE increases from $0.0061$ on the source holdout to $2.42$ on M4Raw, and median log-NMSE shifts by approximately $6.0$ natural-log units. We therefore treat this result as a frozen-reconstructor stress-test failure rather than successful deployment transfer. Consistent with the label-location shift, all source-calibrated point predictors incur multi-unit target RMSE in Table~\ref{tab:target_rmse}. RM-RMS reverses similarly, while CQR remains positively oriented at $r=0.680\pm0.067$ and $\rho_s=0.718\pm0.066$. Supplementary Sec. S7 gives the operating-range and target-cache audits.

\begin{table*}[t]
    \centering
    \footnotesize
    \caption{Matched M4Raw scan-level zero-shot transfer. Values are signed correlations with log-NMSE. Negative values indicate a reversed source-trained readout. Uncertainty follows Table~\ref{tab:source_matched}.}
    \label{tab:target_matched}
    \setlength{\tabcolsep}{4.1pt}
    \begin{tabular}{lcccccc}
        \toprule
        & \multicolumn{2}{c}{XPDNet} & \multicolumn{2}{c}{VarNet} & \multicolumn{2}{c}{PromptMR+} \\
        \cmidrule(lr){2-3}\cmidrule(lr){4-5}\cmidrule(lr){6-7}
        Method & $r$ & $\rho_s$ & $r$ & $\rho_s$ & $r$ & $\rho_s$ \\
        \midrule
        \textbf{Trajectory} & $+0.630\pm0.086$ & $+0.662\pm0.090$ & $-0.935\pm0.020$ & $-0.924\pm0.030$ & $+0.856\pm0.043$ & $+0.852\pm0.052$ \\
        Hu-style & $+0.651\pm0.099$ & $+0.647\pm0.113$ & $-0.746\pm0.072$ & $-0.731\pm0.092$ & $+0.648\pm0.116$ & $+0.702\pm0.109$ \\
        ResNet$^\dagger$ & $+0.344\pm0.113$ & $+0.389\pm0.129$ & $+0.308\pm0.152$ & $+0.329\pm0.161$ & $+0.292\pm0.207$ & $+0.313\pm0.191$ \\
        CQR & $+0.298\pm0.222$ & $+0.196\pm0.187$ & $+0.680\pm0.067$ & $+0.718\pm0.066$ & $+0.422\pm0.184$ & $+0.298\pm0.184$ \\
        RM-RMS & $+0.289\pm0.126$ & $+0.270\pm0.161$ & $-0.917\pm0.020$ & $-0.922\pm0.025$ & $+0.096\pm0.139$ & $+0.092\pm0.174$ \\
        \bottomrule
    \end{tabular}
\end{table*}

Signed correlation and source-scale RMSE expose separate transfer properties. PromptMR+ trajectory preserves target ordering at $r=0.856\pm0.043$, while its zero-shot RMSE is $0.459\pm0.018$. Hu-style has weaker correlation at $0.648\pm0.116$ and lower RMSE at $0.165\pm0.023$. XPDNet shows a third calibration ordering. ResNet has the lowest zero-shot RMSE at $0.192\pm0.025$, trajectory follows at $0.225\pm0.020$, and Hu-style reaches $0.417\pm0.017$, despite trajectory and Hu-style having similar target correlation. The source trajectory readout therefore requires target recalibration for absolute log-risk even when its ordering transfers. VarNet combines reversed ordering with large source-scale RMSE. Table~\ref{tab:target_rmse} reports these calibration results without treating them as ranking metrics.

\begin{table}[t]
    \centering
    \small
    \caption{M4Raw zero-shot log-NMSE RMSE for matched point predictors. Values are estimate $\pm$ patient-cluster bootstrap SD.}
    \label{tab:target_rmse}
    \setlength{\tabcolsep}{3.6pt}
    \begin{tabular}{lccc}
        \toprule
        Backbone & Trajectory & Hu-style & ResNet$^\dagger$ \\
        \midrule
        XPDNet & $0.225\pm0.020$ & $0.417\pm0.017$ & $0.192\pm0.025$ \\
        VarNet & $6.538\pm0.215$ & $4.403\pm0.054$ & $4.317\pm0.033$ \\
        PromptMR+ & $0.459\pm0.018$ & $0.165\pm0.023$ & $0.989\pm0.025$ \\
        \bottomrule
    \end{tabular}
\end{table}

Trajectory risk can become informative before reconstruction is complete. On XPDNet, a source-only predictor using the first 12 of 25 iterations reaches target $r=0.622$, compared with $r=0.630$ for the full path. Source correlation rises from $0.700$ to $0.804$. Supplementary Table S10 gives the complete prefix comparison.

\begin{figure*}[t]
    \centering
    \includegraphics[width=0.98\textwidth]{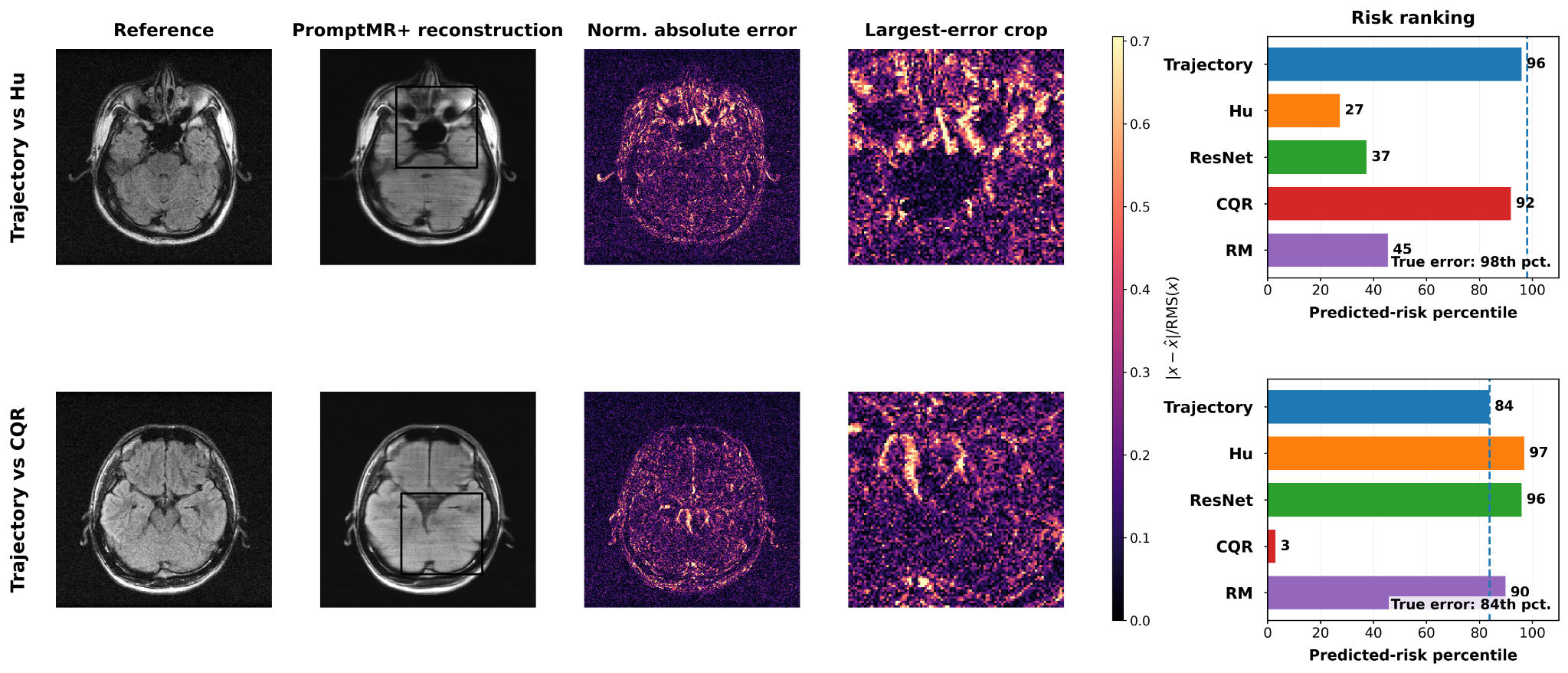}
    \caption{PromptMR+ M4Raw cases prioritized by trajectory risk when individual baselines under-rank the error. Hu denotes the Hu-style comparator.}
    \label{fig:qualitative}
\end{figure*}

Figure~\ref{fig:qualitative} shows two high-error PromptMR+ cases with different comparator failures. In the first case, true error is at the 98th percentile and trajectory risk is at the 96th, while Hu-style, ResNet, and RM-RMS assign percentiles 27, 37, and 45. CQR assigns the 92nd percentile. In the second case, trajectory risk matches the 84th true-error percentile, while CQR assigns the 3rd percentile and the other endpoint scores rank the case highly. These examples illustrate heterogeneous score failures. Population-level comparisons remain those in Table~\ref{tab:target_matched}.

\subsection{Complementarity, Initialization, and Physics-Side Evidence}
\label{sec:complementarity}

The trajectory and Hu-style predictions are related, but their overlap depends on the backbone. On PromptMR+, their target Pearson correlation is $0.712$. The partial association between trajectory and log-NMSE after conditioning on Hu-style remains $0.739\pm0.111$, while the partial association of Hu-style after conditioning on trajectory is $0.104\pm0.205$. The corresponding 95\% intervals are $[0.430,0.868]$ and $[-0.291,0.499]$. Most conditional signal in this experiment is carried by the trajectory.

For XPDNet, the trajectory-Hu-style prediction correlation is $0.421$. Both scores retain target association after conditioning on the other. Partial Pearson correlations are $0.517$ with CI $[0.341,0.679]$ for trajectory given Hu-style and $0.548$ with CI $[0.316,0.729]$ for Hu-style given trajectory. Source-only meta-out-of-fold fusion increases target Pearson correlation from $0.630$ to $0.746$ and Spearman correlation from $0.662$ to $0.745$. Paired improvement intervals are $[0.035,0.195]$ and $[0.0001,0.165]$. These results establish complementarity without requiring target fitting.

VarNet also contains conditional structure, with partial Pearson $-0.863\pm0.052$ for trajectory given Hu-style and $-0.320\pm0.113$ for Hu-style given trajectory. Both retain the reversed orientation. This is evidence about retained structure rather than successful prediction.

We next test whether trajectory risk reflects difficulty visible at initialization. On PromptMR+, the learned initialization-only predictor is much weaker than trajectory on source data ($r=0.218$ versus $r=0.906$), and conditioning on it leaves trajectory association nearly unchanged at partial $r=0.901\pm0.017$. Conversely, the initialization-only score given trajectory has essentially no residual source association ($r=-0.000\pm0.074$). Under M4Raw shift, trajectory and initialization-only prediction reach $r=0.856$ and $r=0.836$, respectively. Both retain substantial conditional association with error: partial $r=0.644\pm0.075$ for trajectory given initialization and $r=0.581\pm0.123$ for initialization given trajectory. Thus, the source trajectory largely subsumes the learned initialization control, while under target shift the two signals become complementary. For VarNet, conditioning on $q_0$ leaves the reversed target association nearly unchanged at $r=-0.926\pm0.022$, while $q_0$ given trajectory reaches $r=0.027\pm0.164$. Supplementary Table~S8 gives the full bidirectional analysis.

The final data-consistency residual provides a separate diagnostic. It is weak on VarNet source data at $r=0.214$, then reaches $r=0.974$ on M4Raw. The same target score remains weak for PromptMR+ at $r=0.149$. Adding cheap controls to the source trajectory readout leaves the VarNet correlation negative at $r=-0.919$. To test whether the transfer pattern is an artifact of trajectory-scale normalization, we retain only ten descriptor components invariant to uniform trajectory scaling. PromptMR+ remains positively associated on M4Raw at $r=0.844$, while the VarNet reversal becomes slightly stronger at $r=-0.958$. Supplementary Tables S9 and S12 give the complete scale and cheap-control results.

Supplementary Sec. S4 gives the full endpoint-conditional matrix. The sign remains a property of the deployed readout. Magnitude alone is not used to label the VarNet score successful.

\subsection{Signal Content and Readout Repair}
\label{sec:mechanism}

\textbf{Discovery and selective risk.}
The earlier single-coil eight-cascade discovery model tests the path-geometry hypothesis before the shared descriptor was defined. On a fresh 100-volume cohort, its endpoint-amplitude model reaches $r=0.8930$, $\rho_s=0.9521$, and RMSE $0.3873$. The frozen discovery-model trajectory descriptor reaches $r=0.9635$, $\rho_s=0.9718$, and RMSE $0.2519$. The paired trajectory-minus-amplitude RMSE difference is $-0.1359$ with 95\% CI $[-0.1987,-0.0794]$. The Spearman difference is $0.0202$ with CI $[-0.0049,0.0528]$. This result supports improved calibrated error prediction without resolving a rank-correlation advantage on that cohort. This model is used only for discovery and mask decomposition rather than for the shared-descriptor generalization claim. Supplementary Sec. S8 gives the full analysis.

\textbf{Repeated masks.}
The repeated-mask collection reconstructs 199 physical volumes with eight independent masks, giving 1592 volume-mask pairs. Across this full descriptive collection, total log-risk variance is $0.7281$, while mean within-volume variance is $0.00457$, only $0.63\%$ of the total. The mean maximum-to-minimum NMSE ratio over masks is $1.21$. Inferential associations are evaluated on the previously frozen 100-volume test subset of 800 reconstructions. Trajectory strongly ranks persistent between-volume differences, with $\rho_s=0.9763\pm0.0056$. Its centered within-volume association is small and negative at $\rho_s=-0.1205\pm0.0405$. After removing realized mask-density effects, the residual association is small and positive at $\rho_s=0.1419\pm0.0412$. Endpoint amplitude reaches within-volume $\rho_s=0.6704\pm0.0347$, and the cluster-LOO mask-density control reaches $0.6929\pm0.0168$. Under this repeated-mask protocol, trajectory primarily ranks persistent case-level difficulty rather than fine changes induced by nearby sampling-mask realizations.

\begin{table}[t]
    \centering
    \footnotesize
    \caption{Repeated-mask Spearman associations for the discovery model on the frozen 100-volume test subset of 800 volume-mask reconstructions. Values are estimate $\pm$ source-volume-cluster bootstrap SD from 5,000 replicates. Between averages the eight masks within each volume. Within centers predictions and targets within physical volume. Partial correlation additionally removes realized mask density. The mask-density control uses the canonical cluster-LOO mask-fraction predictor.}
    \label{tab:multimask}
    \setlength{\tabcolsep}{2.5pt}
    \begin{tabular}{lcccc}
        \toprule
        Score & Pooled & Between & Within & Partial \\
        \midrule
        Amplitude & \shortstack{$0.9505$\\$\pm0.0128$} & \shortstack{$0.9507$\\$\pm0.0139$} & \shortstack{$0.6704$\\$\pm0.0347$} & \shortstack{$0.4066$\\$\pm0.0491$} \\
        Trajectory & \shortstack{$0.9660$\\$\pm0.0062$} & \shortstack{$0.9763$\\$\pm0.0056$} & \shortstack{$-0.1205$\\$\pm0.0405$} & \shortstack{$0.1419$\\$\pm0.0412$} \\
        Mask density & -- & -- & \shortstack{$0.6929$\\$\pm0.0168$} & -- \\
        \bottomrule
    \end{tabular}
\end{table}

The matched target associations are also not purely between-patient effects. After centering score and log-NMSE within patient, Pearson correlation remains $-0.924$ for VarNet, $0.523$ for PromptMR+, and $0.580$ for XPDNet. These analyses probe different granularities. The repeated-mask test changes only the sampling realization of the same physical volume and uses the discovery-model descriptor, whereas the matched within-patient analysis retains scan- and acquisition-level variation within a patient and uses the shared trajectory descriptor. Supplementary Table S11 gives the full within-patient results.

\textbf{Severe shift and calibration.}
The representation-readout separation is visible directly in matched VarNet. Across patient-held-out splits, fitting only offset and slope from four M4Raw patients changes mean held-out Pearson correlation to $0.935\pm0.009$ and reduces RMSE from $0.231\pm0.027$ for a target constant to $0.092\pm0.017$. Similar results are obtained with 8 and 16 patients. This is a target-adaptation diagnostic rather than zero-shot success. Supplementary Table S14 gives the repeated-split results.

A separate stress test trains a frozen PromptMR trajectory coordinate on fastMRI knee and evaluates M4Raw. The pooled target slope reverses in the adaptation cohort at $-0.916$, with 95\% CI $[-1.254,-0.577]$, and in a disjoint confirmation cohort at $-1.383$, with CI $[-1.641,-1.127]$. Patient-and-acquisition adjusted slopes remain negative at $-0.546$ and $-0.323$. Supplementary Table S18 gives the strict models and all intervals.

Only the offset and slope in Eq.~\ref{eq:calibration} are fitted on 4, 8, or 16 adaptation patients. Evaluation uses a disjoint 25-patient confirmation cohort. RMSE reductions relative to a target constant range from $6.3\%$ to $7.3\%$. Paired intervals favor trajectory over calibrated amplitude for 4 and 8 patients, while the 16-patient interval includes zero. Supplementary Tables S18--S19 give the full slopes, calibration values, and intervals.

Taken together, the controls support a distinction between representation informativeness and readout transfer. Initialization conditioning and scale-invariant
features preserve substantial trajectory association, while severe
shift can reverse the source-trained readout without eliminating the
underlying target ordering. The repeated-mask and within-patient
analyses further show that this ordering is primarily case-level but is
not explained solely by patient identity. These results motivate
treating trajectory risk as a representation whose usefulness depends
on both the information encoded by the trajectory and the orientation
of the deployed scalar readout.
\section{Discussion}
\label{sec:discussion}

The matched experiment makes architecture the main varying factor by holding the cohorts, slices, mask semantics, risk target, and feature functions fixed. Source trajectory correlations occupy a narrow high range, while target correlations span $-0.935$ to $0.856$. Image-space solver geometry contains substantial risk structure across the evaluated reconstructors, while the source coefficient vector does not define an invariant orientation under domain shift.

The comparator matrix clarifies the role of this modality. PromptMR+ trajectories retain most conditional association after controlling for Hu-style. XPDNet trajectories and Hu-style predictions both retain conditional association, and their source-only fusion improves target performance. VarNet trajectory and RM-RMS reverse, while CQR and the data-consistency residual remain positively associated. Solver dynamics therefore provide distinct evidence whose relative value depends on the reconstruction backbone and shift.

The initialization and repeated-mask controls refine what the trajectory represents. On source data, the learned initialization-only control has little independent association once the trajectory is known. Under target shift, initialization can become strongly informative, consistent with increased input-visible case difficulty, while the trajectory remains substantially and independently associated with error. The repeated-mask experiment further shows that, under that protocol, most trajectory signal reflects persistent differences between cases rather than fine variation among nearby sampling masks. We therefore interpret inference trajectories as a model-internal representation of case-level reconstruction difficulty that can retain information beyond a learned initialization-only score, rather than as a universally dynamics-exclusive measure of error.

Patient centering does not remove acquisition-level structure. In a separate earlier XPDNet transfer analysis, patient-and-acquisition adjustment reduced the pooled target slope from $1.560$ to $0.0935$ in Supplementary Sec.~S9. The matched within-patient correlations show that the associations are not explained solely by differences between patients, although acquisition-level variation remains. The VarNet reversal also persists under the scale-invariant and initialization controls, but its origin remains unclear.

Correlation and RMSE answer different deployment questions. Positive rank association can support triage after monotone target calibration, while absolute risk thresholds require correct offset and scale. PromptMR+ illustrates this distinction through high zero-shot correlation and weaker source-scale RMSE. VarNet fails signed ordering and calibration. The matched VarNet and independent PromptMR experiments show that two target parameters can recover calibrated prediction when the frozen coordinate remains ordered. Such repair requires labeled target data, since a readout reversal cannot be identified from the source score alone. Their different split designs are reported in Supplementary Secs. S7 and S9 and do not support a general sample-complexity claim.

The descriptor reuses one deterministic reconstruction and applies a 16-feature linear readout. Early prediction has been demonstrated only for XPDNet, and wall-clock costs were not compared across the differently instrumented methods. The PromptMR+ scalar ResNet also shows seed sensitivity, so endpoint conclusions rely on the broader comparator and conditional analyses.

The evaluation covers three backbones, one source dataset, one M4Raw FLAIR cohort, fixed nominal acceleration, and log-NMSE. The repeated-mask analysis uses the earlier discovery-model descriptor and nearby masks. Future work should test additional anatomies, field strengths, accelerations, sampling patterns, and clinically weighted targets. Prospective studies should evaluate review decisions, instrumentation cost, and target-efficient orientation diagnosis and calibration.

\clearpage

{
    \small
    \bibliographystyle{ieeenat_fullname}
    \bibliography{references}
}
\clearpage
\appendix
\section*{Appendix}
This supplement gives the exact trajectory descriptors, matched evaluation details, bootstrap intervals, comparator definitions, initialization-normalization controls, early-prefix results, operating-range audits, and the discovery and stress-test experiments. Single-method uncertainty for XPDNet, VarNet, and PromptMR+ uses 10,000 patient-cluster bootstrap replicates. The reported SDs describe bootstrap sampling variation rather than variation across training runs. Paired comparisons use matched patient resamples.

\section{Trajectory Descriptors}
\label{app:features}

\subsection{Generic 16-Dimensional Descriptor}

Let $q_0,\ldots,q_T$ be projected image states, $\Delta_t=q_t-q_{t-1}$, $s_0=\rms(q_0)+\epsilon$, and $d_t=\rms(\Delta_t)/s_0$. The implementation uses raw iteration indices $u_t=t$ for $t=1,\ldots,T$ and sets $h=\lfloor T/3\rfloor$. Early and late step windows contain $d_1,\ldots,d_h$ and $d_{T-h+1},\ldots,d_T$. Linear slopes are ordinary least-squares slopes against the corresponding raw indices. Standard deviations use the population convention.

For endpoint distances, $a_t=\rms(q_T-q_t)/(\rms(q_T)+\epsilon)$ for $t=0,\ldots,T$. The stored endpoint-area features are arithmetic means. The overall value is $(T+1)^{-1}\sum_{t=0}^{T}a_t$, and the late value is $(h+1)^{-1}\sum_{t=T-h}^{T}a_t$. Consecutive-update cosine is computed by direct division when both norms are positive and is set to zero when either norm is zero. Its late mean uses the final $h$ cosine values. The numerical constant is $\epsilon=10^{-12}$.

\begin{table*}[t]
    \centering
    \footnotesize
    \caption{Shared 16-dimensional descriptor used without definition changes for XPDNet, VarNet, and PromptMR+. Stored names are retained from the frozen helper.}
    \label{tab:features}
    \setlength{\tabcolsep}{4.5pt}
    \begin{tabular}{p{43mm}p{48mm}p{70mm}}
        \toprule
        Stored name & Definition & Role \\
        \midrule
        \nolinkurl{trajectory_path_length} & $\sum_{t=1}^{T}d_t$ & Total normalized solver motion \\
        \nolinkurl{trajectory_endpoint_displacement} & $\rms(q_T-q_0)/s_0$ & Net motion from initialization to output \\
        \nolinkurl{trajectory_path_efficiency} & endpoint displacement divided by path length & Directness of the image path \\
        \nolinkurl{trajectory_step_mean} & mean of $\{d_t\}$ & Typical update magnitude \\
        \nolinkurl{trajectory_step_std} & population SD of $\{d_t\}$ & Variation in update magnitude \\
        \nolinkurl{trajectory_step_max} & maximum of $\{d_t\}$ & Largest update \\
        \nolinkurl{trajectory_step_final} & $d_T$ & Motion at the endpoint \\
        \nolinkurl{trajectory_step_late_over_early} & mean final-third step divided by mean first-third step & Temporal concentration of motion \\
        \nolinkurl{trajectory_log_step_slope} & slope of $\log(d_t+\epsilon)$ on $t=1,\ldots,T$ & Global update decay or growth \\
        \nolinkurl{trajectory_late_log_step_slope} & slope over $t=T-h+1,\ldots,T$ & Late update decay or growth \\
        \nolinkurl{trajectory_endpoint_distance_auc} & arithmetic mean of $a_0,\ldots,a_T$ & Overall rate of endpoint approach \\
        \nolinkurl{trajectory_late_endpoint_distance_auc} & arithmetic mean of $a_{T-h},\ldots,a_T$ & Residual motion near the endpoint \\
        \nolinkurl{trajectory_endpoint_monotone_fraction} & fraction of transitions with $a_t\leq a_{t-1}$ & Consistency of endpoint approach \\
        \nolinkurl{trajectory_consecutive_delta_cos_mean} & mean consecutive-update cosine & Average directional agreement \\
        \nolinkurl{trajectory_consecutive_delta_cos_std} & SD of consecutive-update cosine & Variation in directional agreement \\
        \nolinkurl{trajectory_late_delta_cos_mean} & mean of final $h$ consecutive-update cosines & Late directional agreement \\
        \bottomrule
    \end{tabular}
\end{table*}

XPDNet exposes 25 projected primal-dual updates. VarNet exposes 12 sensitivity-reduced cascade outputs. PromptMR+ exposes an initialization and 12 central-slice image outputs from its five-slice, history-aware recurrence. Each trajectory is aligned to the reconstruction target before feature extraction. The same 16 functions are then applied to all three paths. All window definitions are expressed relative to the available trajectory length, so the same feature functions are applied without changing their definitions across the 25-update XPDNet path and the 12-stage VarNet and PromptMR+ paths.

Ten descriptor components are independent of the $q_0$-based normalization $s_0$ after algebraic cancellation: path efficiency, the late-to-early step ratio, both log-step slopes, the three endpoint-approach quantities, and the three cosine summaries. We use these components as an $s_0$-independent diagnostic of whether the choice of initialization normalization drives the transfer behavior. The ten stored features are \nolinkurl{trajectory_path_efficiency}, \nolinkurl{trajectory_step_late_over_early}, \nolinkurl{trajectory_log_step_slope}, \nolinkurl{trajectory_late_log_step_slope}, \nolinkurl{trajectory_endpoint_distance_auc}, \nolinkurl{trajectory_late_endpoint_distance_auc}, \nolinkurl{trajectory_endpoint_monotone_fraction}, \nolinkurl{trajectory_consecutive_delta_cos_mean}, \nolinkurl{trajectory_consecutive_delta_cos_std}, and \nolinkurl{trajectory_late_delta_cos_mean}.

\subsection{Frozen Discovery-Model Descriptor}

The earlier single-coil eight-cascade discovery model uses a separate descriptor fixed before the matched three-backbone study. It contains eight cascade states $q_0,\ldots,q_7$. For $t=1,\ldots,7$, define
\begin{align}
    d_t^{\mathrm{D}}
    &=\frac{\rms(q_t-q_{t-1})}{\rms(q_t)+\epsilon},
    &L^{\mathrm{D}}&=\sum_{t=1}^{7}d_t^{\mathrm{D}},\\
    D^{\mathrm{D}}
    &=\frac{\rms(q_7-q_0)}{\rms(q_7)+\epsilon}.
\end{align}
Path efficiency is $\eta^{\mathrm{D}}=D^{\mathrm{D}}/(L^{\mathrm{D}}+\epsilon)$. Late step mean and maximum use $d_5^{\mathrm{D}},d_6^{\mathrm{D}},d_7^{\mathrm{D}}$. The descriptor also retains the last two consecutive-step cosines $c_6,c_7$ and the final change $\delta_D$ in the model's auxiliary degradation-probability tensor.

The discovery model evaluates a fixed set $\mathcal{S}$ of four slices per volume. For a slice-level quantity $f$, let $\mu_{\mathcal{S}}(f)$ and $\sigma_{\mathcal{S}}(f)$ be its mean and population SD. The frozen vector is
\begin{equation}
\begin{split}
\phi_{\mathrm{disc}}=[
&\mu(L^{\mathrm{D}}),\sigma(L^{\mathrm{D}}),
\mu(\eta^{\mathrm{D}}),\sigma(\eta^{\mathrm{D}}),
\mu(D^{\mathrm{D}}),\sigma(D^{\mathrm{D}}),\\
&\mu(\mu_{\mathrm{late}}^{\mathrm{D}}),
\sigma(\mu_{\mathrm{late}}^{\mathrm{D}}),
\mu(m_{\mathrm{late}}^{\mathrm{D}}),
\sigma(m_{\mathrm{late}}^{\mathrm{D}}),\\
&\mu(d_5^{\mathrm{D}}),\mu(d_6^{\mathrm{D}}),
\mu(d_7^{\mathrm{D}}),\mu(c_6),\mu(c_7),\mu(\delta_D)].
\end{split}
\end{equation}
The model supports only the discovery and repeated-mask analyses. It is excluded from the shared-descriptor backbone-generalization claim. Literal feature reuse is established by the matched XPDNet, VarNet, and PromptMR+ experiment.

\section{Matched Protocol and Comparator Definitions}
\label{app:protocol}

The source cohort contains 3676 fastMRI-brain slices. Source fitting uses 2944 rows from 736 patients, and held-out evaluation uses 732 rows from 183 patients. Four central slices remain grouped within each volume. The M4Raw FLAIR target contains 400 slices from 100 scans and 25 patients. Four selected slices are averaged to scan level before target correlation and patient resampling.

Feature normalization, ridge selection, and all comparator fitting use source data. No M4 label is used for fitting, model selection, orientation correction, or calibration in the matched zero-shot study. The bootstrap uses 10,000 replicates with seed 20260828. The source independent unit is the patient, with 183 held-out patients and 732 slices. The M4Raw independent unit is the patient, with 25 patients. M4Raw predictions are first aggregated to 100 FLAIR-scan scores before resampling. Percentile confidence intervals use the 2.5th and 97.5th percentiles. Paired comparisons apply the same patient resample to both methods.

Trajectory is the source-only ridge predictor of log-NMSE from the generic descriptor. Our Hu-style comparator adapts the two-step error-prediction idea of Hu et al. to a final-reconstruction input \cite{hu2021predict}. ResNet is a learned final-image scalar log-NMSE predictor. CQR is normalized mean prediction-interval width \cite{angelopoulos2022im2im,romano2019cqr}. RM-RMS is normalized RMS residual-magnitude interval width following image-to-image risk control \cite{angelopoulos2022im2im}. The XPDNet RMS-normalized CQR width is retained as a secondary score in Sec.~\ref{app:controls}. The primary CQR result always uses mean-normalized width.

For the VarNet and PromptMR+ transfer analyses, we additionally evaluate endpoint amplitude and the final sampled-space data-consistency residual as inexpensive diagnostics.

RMSE is meaningful for calibrated point predictors such as trajectory, Hu-style, and ResNet. CQR and RM-RMS are interval-width risk scores, so the main comparison reports their signed correlation with log-NMSE. Fusion is analyzed separately because the XPDNet model uses a meta-out-of-fold construction, while the VarNet and PromptMR+ stacks are less strictly cross-fitted. Complementarity claims for VarNet and PromptMR+ therefore use partial correlations.

\section{Full Matched Results}
\label{app:matched}

\begin{table*}[t]
    \centering
    \footnotesize
    \caption{VarNet matched results with patient-cluster uncertainty. Every metric is estimate $\pm$ bootstrap SD followed by its 95\% percentile CI. Target values are signed zero-shot correlations.}
    \label{tab:varnet_full}
    \setlength{\tabcolsep}{3.5pt}
    \begin{tabular}{llcccc}
        \toprule
        Split & Method & $r\pm\mathrm{SD}$ & 95\% CI & $\rho_s\pm\mathrm{SD}$ & 95\% CI \\
        \midrule
        Source & Trajectory & $+0.849\pm0.024$ & $[0.796,0.891]$ & $+0.834\pm0.022$ & $[0.784,0.871]$ \\
        & Hu-style & $+0.757\pm0.029$ & $[0.694,0.808]$ & $+0.734\pm0.033$ & $[0.661,0.791]$ \\
        & ResNet & $+0.588\pm0.050$ & $[0.481,0.675]$ & $+0.551\pm0.051$ & $[0.442,0.639]$ \\
        & CQR & $+0.819\pm0.022$ & $[0.773,0.858]$ & $+0.823\pm0.024$ & $[0.769,0.863]$ \\
        & RM-RMS & $-0.246\pm0.065$ & $[-0.374,-0.120]$ & $-0.282\pm0.068$ & $[-0.414,-0.146]$ \\
        & Trajectory+Hu-style & $+0.895\pm0.017$ & $[0.858,0.924]$ & $+0.883\pm0.018$ & $[0.843,0.912]$ \\
        \midrule
        M4Raw & Trajectory & $-0.935\pm0.020$ & $[-0.966,-0.888]$ & $-0.924\pm0.030$ & $[-0.960,-0.845]$ \\
        & Hu-style & $-0.746\pm0.072$ & $[-0.861,-0.579]$ & $-0.731\pm0.092$ & $[-0.863,-0.505]$ \\
        & ResNet & $+0.308\pm0.152$ & $[-0.007,0.581]$ & $+0.329\pm0.161$ & $[-0.018,0.600]$ \\
        & CQR & $+0.680\pm0.067$ & $[0.531,0.791]$ & $+0.718\pm0.066$ & $[0.547,0.806]$ \\
        & RM-RMS & $-0.917\pm0.020$ & $[-0.949,-0.871]$ & $-0.922\pm0.025$ & $[-0.950,-0.852]$ \\
        & Trajectory+Hu-style & $-0.939\pm0.018$ & $[-0.969,-0.898]$ & $-0.932\pm0.025$ & $[-0.962,-0.867]$ \\
        \bottomrule
    \end{tabular}
\end{table*}

\begin{table*}[t]
    \centering
    \footnotesize
    \caption{PromptMR+ matched results with patient-cluster uncertainty. Formatting follows Table~\ref{tab:varnet_full}.}
    \label{tab:promptmr_full}
    \setlength{\tabcolsep}{3.5pt}
    \begin{tabular}{llcccc}
        \toprule
        Split & Method & $r\pm\mathrm{SD}$ & 95\% CI & $\rho_s\pm\mathrm{SD}$ & 95\% CI \\
        \midrule
        Source & Trajectory & $+0.906\pm0.016$ & $[0.870,0.935]$ & $+0.895\pm0.017$ & $[0.854,0.923]$ \\
        & Hu-style & $+0.616\pm0.039$ & $[0.531,0.686]$ & $+0.563\pm0.045$ & $[0.467,0.644]$ \\
        & ResNet & $+0.187\pm0.052$ & $[0.083,0.284]$ & $+0.136\pm0.061$ & $[0.014,0.252]$ \\
        & CQR & $+0.855\pm0.018$ & $[0.817,0.887]$ & $+0.860\pm0.019$ & $[0.816,0.892]$ \\
        & RM-RMS & $-0.115\pm0.054$ & $[-0.220,-0.006]$ & $-0.182\pm0.065$ & $[-0.305,-0.052]$ \\
        & Trajectory+Hu-style & $+0.915\pm0.014$ & $[0.884,0.940]$ & $+0.904\pm0.015$ & $[0.869,0.928]$ \\
        \midrule
        M4Raw & Trajectory & $+0.856\pm0.043$ & $[0.752,0.917]$ & $+0.852\pm0.052$ & $[0.714,0.914]$ \\
        & Hu-style & $+0.648\pm0.116$ & $[0.387,0.837]$ & $+0.702\pm0.109$ & $[0.425,0.847]$ \\
        & ResNet & $+0.292\pm0.207$ & $[-0.149,0.666]$ & $+0.313\pm0.191$ & $[-0.092,0.647]$ \\
        & CQR & $+0.422\pm0.184$ & $[-0.048,0.683]$ & $+0.298\pm0.184$ & $[-0.110,0.606]$ \\
        & RM-RMS & $+0.096\pm0.139$ & $[-0.190,0.359]$ & $+0.092\pm0.174$ & $[-0.246,0.434]$ \\
        & Trajectory+Hu-style & $+0.858\pm0.040$ & $[0.761,0.916]$ & $+0.863\pm0.049$ & $[0.731,0.920]$ \\
        \bottomrule
    \end{tabular}
\end{table*}

\begin{table*}[t]
    \centering
    \footnotesize
    \caption{XPDNet matched results with patient-cluster uncertainty. Every metric is estimate $\pm$ bootstrap SD followed by its 95\% percentile CI. Target values are signed zero-shot correlations. CQR uses normalized mean interval width.}
    \label{tab:xpd_full}
    \setlength{\tabcolsep}{3.5pt}
    \begin{tabular}{llcccc}
        \toprule
        Split & Method & $r\pm\mathrm{SD}$ & 95\% CI & $\rho_s\pm\mathrm{SD}$ & 95\% CI \\
        \midrule
        Source & Trajectory & $+0.804\pm0.027$ & $[0.748,0.852]$ & $+0.807\pm0.026$ & $[0.749,0.850]$ \\
        & Hu-style & $+0.805\pm0.028$ & $[0.742,0.854]$ & $+0.774\pm0.030$ & $[0.707,0.827]$ \\
        & ResNet & $+0.911\pm0.014$ & $[0.880,0.936]$ & $+0.900\pm0.015$ & $[0.867,0.924]$ \\
        & CQR & $+0.790\pm0.031$ & $[0.725,0.844]$ & $+0.786\pm0.028$ & $[0.724,0.833]$ \\
        & RM-RMS & $-0.023\pm0.062$ & $[-0.145,0.097]$ & $-0.022\pm0.070$ & $[-0.159,0.115]$ \\
        \midrule
        M4Raw & Trajectory & $+0.630\pm0.086$ & $[0.440,0.775]$ & $+0.662\pm0.090$ & $[0.447,0.800]$ \\
        & Hu-style & $+0.651\pm0.099$ & $[0.423,0.808]$ & $+0.647\pm0.113$ & $[0.367,0.801]$ \\
        & ResNet & $+0.344\pm0.113$ & $[0.122,0.561]$ & $+0.389\pm0.129$ & $[0.094,0.593]$ \\
        & CQR & $+0.298\pm0.222$ & $[-0.195,0.650]$ & $+0.196\pm0.187$ & $[-0.197,0.529]$ \\
        & RM-RMS & $+0.289\pm0.126$ & $[0.016,0.506]$ & $+0.270\pm0.161$ & $[-0.077,0.549]$ \\
        \bottomrule
    \end{tabular}
\end{table*}

\begin{table*}[t]
    \centering
    \footnotesize
    \caption{RMSE of matched point predictors. Each completed value is estimate $\pm$ patient-cluster bootstrap SD with 95\% CI. Lower is better. Large VarNet target RMSEs reflect failed source calibration and orientation under zero-shot transfer.}
    \label{tab:rmse_full}
    \setlength{\tabcolsep}{4pt}
    \begin{tabular}{lllcc}
        \toprule
        Backbone & Split & Method & RMSE $\pm$ SD & 95\% CI \\
        \midrule
        XPDNet & Source & Trajectory & $0.231\pm0.014$ & $[0.204,0.258]$ \\
        & & Hu-style & $0.228\pm0.011$ & $[0.208,0.250]$ \\
        & & ResNet & $0.188\pm0.007$ & $[0.174,0.202]$ \\
        & M4Raw & Trajectory & $0.225\pm0.020$ & $[0.184,0.262]$ \\
        & & Hu-style & $0.417\pm0.017$ & $[0.383,0.450]$ \\
        & & ResNet & $0.192\pm0.025$ & $[0.144,0.240]$ \\
        \midrule
        VarNet & Source & Trajectory & $0.193\pm0.011$ & $[0.172,0.215]$ \\
        & & Hu-style & $0.240\pm0.010$ & $[0.220,0.260]$ \\
        & & ResNet & $0.324\pm0.014$ & $[0.296,0.353]$ \\
        & M4Raw & Trajectory & $6.538\pm0.215$ & $[6.116,6.952]$ \\
        & & Hu-style & $4.403\pm0.054$ & $[4.296,4.507]$ \\
        & & ResNet & $4.317\pm0.033$ & $[4.252,4.379]$ \\
        \midrule
        PromptMR+ & Source & Trajectory & $0.169\pm0.010$ & $[0.150,0.189]$ \\
        & & Hu-style & $0.316\pm0.017$ & $[0.285,0.350]$ \\
        & & ResNet & $0.492\pm0.022$ & $[0.450,0.535]$ \\
        & M4Raw & Trajectory & $0.459\pm0.018$ & $[0.423,0.493]$ \\
        & & Hu-style & $0.165\pm0.023$ & $[0.119,0.210]$ \\
        & & ResNet & $0.989\pm0.025$ & $[0.941,1.038]$ \\
        \bottomrule
    \end{tabular}
\end{table*}

XPDNet target correlation and target RMSE give different method orderings. Trajectory and Hu-style have similar target correlations, while their RMSEs are $0.225\pm0.020$ and $0.417\pm0.017$. ResNet has lower target correlation and the smallest RMSE at $0.192\pm0.025$. PromptMR+ shows a related distinction. Its trajectory score preserves ordering at $r=0.856$ with RMSE $0.459$, while Hu-style has weaker correlation at $r=0.648$ and smaller RMSE $0.165$. The trajectory claim in this zero-shot experiment concerns risk ordering and complementary information. Source-to-target calibration quality is reported separately.

\section{Paired Comparisons and Complementarity}
\label{app:paired}

Table~S7 is the endpoint-conditional matrix referenced in the main paper. Here, ``endpoint'' denotes the Hu-style final-reconstruction predictor. For each backbone, the table reports trajectory association after conditioning on the Hu-style score and the reverse conditional association for both Pearson and Spearman measures.

\begin{table*}[t]
    \centering
    \footnotesize
    \caption{Paired M4Raw comparisons. A row labeled trajectory minus comparator uses that direction. VarNet and PromptMR+ report estimate $\pm$ bootstrap SD and 95\% CI. XPDNet reports the point difference and 95\% CI. The XPDNet fusion row uses fusion minus trajectory.}
    \label{tab:paired_full}
    \setlength{\tabcolsep}{3.1pt}
    \begin{tabular}{lllcccc}
        \toprule
        Backbone & Difference & $\Delta r\pm\mathrm{SD}$ & 95\% CI & $\Delta\rho_s\pm\mathrm{SD}$ & 95\% CI \\
        \midrule
        VarNet & Trajectory $-$ Hu-style & $-0.189\pm0.072$ & $[-0.351,-0.068]$ & $-0.193\pm0.093$ & $[-0.408,-0.049]$ \\
        & Trajectory $-$ ResNet & $-1.243\pm0.152$ & $[-1.515,-0.926]$ & $-1.253\pm0.163$ & $[-1.520,-0.891]$ \\
        & Trajectory $-$ CQR & $-1.615\pm0.075$ & $[-1.739,-1.448]$ & $-1.643\pm0.075$ & $[-1.737,-1.442]$ \\
        & Trajectory $-$ RM-RMS & $-0.018\pm0.022$ & $[-0.062,0.025]$ & $-0.002\pm0.029$ & $[-0.058,0.057]$ \\
        \midrule
        PromptMR+ & Trajectory $-$ Hu-style & $+0.209\pm0.118$ & $[0.002,0.457]$ & $+0.149\pm0.111$ & $[-0.014,0.409]$ \\
        & Trajectory $-$ ResNet & $+0.564\pm0.220$ & $[0.147,1.019]$ & $+0.539\pm0.204$ & $[0.164,0.958]$ \\
        & Trajectory $-$ CQR & $+0.434\pm0.161$ & $[0.205,0.838]$ & $+0.553\pm0.160$ & $[0.271,0.895]$ \\
        & Trajectory $-$ RM-RMS & $+0.761\pm0.135$ & $[0.494,1.023]$ & $+0.760\pm0.164$ & $[0.422,1.063]$ \\
        \midrule
        XPDNet & Trajectory $-$ Hu-style & $-0.021$ & $[-0.215,0.199]$ & $+0.014$ & $[-0.205,0.299]$ \\
        & Trajectory $-$ CQR & $+0.332$ & $[0.005,0.817]$ & $+0.466$ & $[0.113,0.862]$ \\
        & Trajectory $-$ RM-RMS & $+0.342$ & $[0.078,0.636]$ & $+0.392$ & $[0.056,0.759]$ \\
        & Fusion $-$ trajectory & $+0.115$ & $[0.035,0.195]$ & $+0.083$ & $[0.0001,0.165]$ \\
        \bottomrule
    \end{tabular}
\end{table*}

The VarNet differences are interpreted in signed predictive space. CQR is a positive zero-shot predictor while trajectory and RM-RMS are strongly reversed. The trajectory and RM-RMS difference intervals include zero. The PromptMR+ paired Pearson estimate narrowly favors trajectory over Hu-style, while the corresponding Spearman interval includes zero. It exceeds the reported ResNet realization, CQR, and RM-RMS in both paired metrics. XPDNet trajectory and Hu-style are statistically competitive, and the clean source-only fusion improves over trajectory alone. For XPDNet ResNet, Table~\ref{tab:rmse_full} gives the calibrated point-prediction comparison and Table~\ref{tab:xpd_full} gives marginal correlation uncertainty.

\begin{table*}[t]
    \centering
    \footnotesize
    \caption{Conditional association with M4Raw log-NMSE after controlling for the other score. VarNet and PromptMR+ report estimate $\pm$ bootstrap SD and 95\% CI. XPDNet reports estimate and 95\% CI.}
    \label{tab:partial_full}
    \setlength{\tabcolsep}{3.4pt}
    \begin{tabular}{lccccc}
        \toprule
        Backbone & Prediction corr. & Partial measure & Trajectory given Hu-style & Hu-style given trajectory \\
        \midrule
        PromptMR+ & $r=0.712$ & Pearson & $0.739\pm0.111$ $[0.430,0.868]$ & $0.104\pm0.205$ $[-0.291,0.499]$ \\
        & $\rho_s=0.725$ & Spearman & $0.698\pm0.122$ $[0.393,0.858]$ & $0.235\pm0.173$ $[-0.134,0.526]$ \\
        VarNet & $r=0.713$ & Pearson & $-0.863\pm0.052$ $[-0.935,-0.731]$ & $-0.320\pm0.113$ $[-0.536,-0.087]$ \\
        & $\rho_s=0.677$ & Spearman & $-0.855\pm0.067$ $[-0.926,-0.666]$ & $-0.375\pm0.122$ $[-0.588,-0.101]$ \\
        XPDNet & $r=0.421$ & Pearson & $0.517$ $[0.341,0.679]$ & $0.548$ $[0.316,0.729]$ \\
        & $\rho_s=0.441$ & Spearman & $0.550$ $[0.343,0.708]$ & $0.529$ $[0.213,0.709]$ \\
        \bottomrule
    \end{tabular}
\end{table*}

PromptMR+ trajectory retains a large conditional association after Hu-style, while the Hu-style conditional interval includes zero. XPDNet retains substantial conditional association in both directions, which is consistent with its fusion improvement. VarNet retains negative conditional associations because both source readouts reverse on the target.

\section{Initialization, Normalization, and Early Prediction}
\label{app:additional_controls}

The initialization-only control uses the same source-trained scalar ResNet architecture as the final-image predictor, with $q_0$ as input. On PromptMR+, raw trajectory and $q_0$ correlations are $0.9061$ and $0.2184$ on source, then $0.8565$ and $0.8357$ on M4Raw. The corresponding source rank correlations are $0.8947$ and $0.2160$, and target rank correlations are $0.8517$ and $0.8018$. VarNet source Pearson correlations are $0.8488$ and $0.6413$. On M4Raw, trajectory and $q_0$ reach Pearson $-0.9347$ and $0.3367$ and Spearman $-0.9242$ and $0.3678$.

\begin{table*}[t]
    \centering
    \scriptsize
    \caption{Conditional association with log-NMSE for trajectory and initialization. Values are estimate $\pm$ patient-cluster bootstrap SD with 95\% percentile CI from 10,000 replicates. Each column residualizes the named score against the other score.}
    \label{tab:q0_full}
    \setlength{\tabcolsep}{4pt}
    \begin{tabular}{lllcc}
        \toprule
        Backbone & Split & Measure & Trajectory given $q_0$ & $q_0$ given trajectory \\
        \midrule
        PromptMR+ & Source & Pearson & $0.9011\pm0.0167$ $[0.8654,0.9299]$ & $-0.0004\pm0.0736$ $[-0.1460,0.1415]$ \\
        & & Spearman & $0.8894\pm0.0178$ $[0.8492,0.9184]$ & $0.0404\pm0.0811$ $[-0.1195,0.1959]$ \\
        & M4Raw & Pearson & $0.6440\pm0.0749$ $[0.4696,0.7613]$ & $0.5807\pm0.1228$ $[0.2971,0.7694]$ \\
        & & Spearman & $0.6663\pm0.0828$ $[0.4658,0.7913]$ & $0.5266\pm0.1548$ $[0.1436,0.7485]$ \\
        \midrule
        VarNet & Source & Pearson & $0.7721\pm0.0327$ $[0.6983,0.8267]$ & $0.3867\pm0.0542$ $[0.2754,0.4887]$ \\
        & & Spearman & $0.7498\pm0.0340$ $[0.6745,0.8070]$ & $0.3203\pm0.0574$ $[0.2036,0.4305]$ \\
        & M4Raw & Pearson & $-0.9261\pm0.0220$ $[-0.9614,-0.8751]$ & $0.0273\pm0.1640$ $[-0.2866,0.3443]$ \\
        & & Spearman & $-0.9123\pm0.0334$ $[-0.9529,-0.8232]$ & $0.0721\pm0.1699$ $[-0.2826,0.3752]$ \\
        \bottomrule
    \end{tabular}
\end{table*}

On the PromptMR+ source holdout, trajectory retains essentially its full association after conditioning on $q_0$, while the learned initialization-only control contributes no detectable residual association. Under M4Raw shift, both coordinates retain conditional information. On VarNet target data, conditioning on $q_0$ preserves the reversed trajectory association, while $q_0$ adds no detectable residual association given trajectory. These controls show substantial trajectory information beyond the learned initialization-only predictor while allowing input-visible difficulty to contribute under shift. This analysis controls for information captured by the learned initialization-only predictor. It does not establish that no other function of $q_0$ could carry additional risk information.

\begin{table}[t]
    \centering
    \footnotesize
    \caption{Initialization-normalization diagnostic. Full uses all 16 components. The $s_0$-independent model uses the ten components whose values do not depend on the $q_0$-based scale normalizer after algebraic cancellation. Ridge regularization is reselected using source-only five-fold patient-level cross-validation. Values are descriptive source-trained point estimates.}
    \label{tab:s0_independent}
    \setlength{\tabcolsep}{3pt}
    \begin{tabular}{llcccc}
        \toprule
        Backbone & Split & \multicolumn{2}{c}{Full} & \multicolumn{2}{c}{$s_0$-independent} \\
        & & $r$ & $\rho_s$ & $r$ & $\rho_s$ \\
        \midrule
        VarNet & Source & $0.8488$ & $0.8342$ & $0.8183$ & $0.7961$ \\
        & M4Raw & $-0.9347$ & $-0.9242$ & $-0.9575$ & $-0.9482$ \\
        PromptMR+ & Source & $0.9061$ & $0.8947$ & $0.8840$ & $0.8825$ \\
        & M4Raw & $0.8565$ & $0.8517$ & $0.8443$ & $0.8572$ \\
        \bottomrule
    \end{tabular}
\end{table}

The main text refers to this restricted subset as the scale-invariant control. More precisely, the diagnostic isolates the ten components that are independent of the $q_0$-based normalizer $s_0$ after algebraic cancellation.

Restricting the readout to the ten components independent of the $q_0$-based normalizer preserves positive PromptMR+ transfer and slightly strengthens the VarNet reversal. Source-only cross-validation selects ridge regularization strengths $0.001$ for VarNet and $0.1$ for PromptMR+. The initialization normalization therefore does not explain the backbone-dependent target orientation.

\begin{table}[t]
    \centering
    \footnotesize
    \caption{XPDNet source-only trajectory prediction using 12 of 25 or all 25 iterations. Target values use the matched M4Raw cohort. RMSE is reported where evaluated.}
    \label{tab:xpd_prefix}
    \setlength{\tabcolsep}{3.2pt}
    \begin{tabular}{lccccc}
        \toprule
        Prefix & Source $r$ & Source $\rho_s$ & Source RMSE & Target $r$ & Target $\rho_s$ \\
        \midrule
        12/25 & $0.7005$ & $0.6961$ & $0.2740$ & $0.6224$ & $0.6228$ \\
        25/25 & $0.8036$ & $0.8068$ & $0.2311$ & $0.6303$ & $0.6618$ \\
        \bottomrule
    \end{tabular}
\end{table}

Most XPDNet target Pearson association is available by iteration 12. Source prediction continues to improve through iteration 25. This provides an early-triage use for trajectory features that endpoint-only scores cannot supply. This timing diagnostic is evaluated on XPDNet only. It does not establish that the same fraction of inference suffices for the other backbones.

\begin{table}[t]
    \centering
    \small
    \caption{M4Raw association after centering score and log-NMSE within patient. Values are descriptive scan-level point estimates.}
    \label{tab:within_patient}
    \setlength{\tabcolsep}{5pt}
    \begin{tabular}{lcc}
        \toprule
        Backbone & Pearson $r$ & Spearman $\rho_s$ \\
        \midrule
        XPDNet & $0.5800$ & $0.5343$ \\
        VarNet & $-0.9236$ & $-0.8508$ \\
        PromptMR+ & $0.5234$ & $0.5004$ \\
        \bottomrule
    \end{tabular}
\end{table}

The centered associations show that the matched target results are not explained only by between-patient differences. They remain descriptive because centering does not isolate a causal patient effect or remove acquisition-level structure. This analysis retains scan- and acquisition-level variation within each patient and uses the shared descriptor. The repeated-mask analysis in Sec. S8 changes only the sampling realization of one physical volume and uses the discovery-model descriptor.

\section{Controls and Score Variants}
\label{app:controls}

For the joint diagnostic, \emph{Trajectory+controls} concatenates the 16 trajectory features with endpoint amplitude, the final data-consistency residual, and mask density. The resulting 19-input vector is mapped by a single source-trained ridge readout using the same source-only fitting protocol.

\begin{table*}[t]
    \centering
    \footnotesize
    \caption{Backbone-specific endpoint and physics-side diagnostics used to interpret the distinct VarNet and PromptMR+ transfer results. Trajectory+controls is a 19-input source-trained joint readout comprising the 16 trajectory features, endpoint amplitude, the final data-consistency residual, and mask density. Values are descriptive point estimates, while primary inference uses the matched tables.}
    \label{tab:controls}
    \setlength{\tabcolsep}{4pt}
    \begin{tabular}{llcccc}
        \toprule
        Backbone & Score & Source $r$ & Source $\rho_s$ & M4Raw $r$ & M4Raw $\rho_s$ \\
        \midrule
        VarNet & Final DC residual & $0.2142$ & $0.1086$ & $0.9741$ & $0.9658$ \\
        & Endpoint amplitude & $-0.0097$ & $-0.0117$ & $0.1504$ & $0.1900$ \\
        & Trajectory+controls & $0.8708$ & $0.8570$ & $-0.9185$ & $-0.8956$ \\
        PromptMR+ & Final DC residual & $-0.1417$ & $-0.1628$ & $0.1488$ & $0.2064$ \\
        & Endpoint amplitude & $-0.0086$ & $-0.0231$ & $0.0835$ & $0.0602$ \\
        & Trajectory+controls & $0.9097$ & $0.8974$ & $0.8459$ & $0.8459$ \\
        \bottomrule
    \end{tabular}
\end{table*}

The VarNet DC residual is weak on source and highly correlated with error after the M4Raw shift. It is therefore a strong deployment-specific target proxy rather than evidence of a universally transferable source readout. The joint cheap-control model does not repair the trajectory reversal. PromptMR+ endpoint and DC controls are weak, and their addition slightly reduces target correlation.

RM-RMS is an uncertainty-width score rather than a supervised point predictor. Its signed association with realized log-NMSE is empirical. Its negative source correlations for VarNet and PromptMR+ indicate that interval width does not monotonically order realized error under this protocol.

For XPDNet, the primary CQR score is normalized mean interval width. It gives source $r=0.7896$, $\rho_s=0.7861$ and target $r=0.2979$, $\rho_s=0.1959$. A secondary RMS-normalized width gives source $r=0.8498$, $\rho_s=0.8334$ and target $r=0.3853$, $\rho_s=0.3205$. The main paper consistently uses the primary mean-width definition. XPDNet RM-RMS uses normalized RMS interval width and gives source $r=-0.0235$, $\rho_s=-0.0222$ and target $r=0.2887$, $\rho_s=0.2701$.

\section{Operating Range and Robustness Diagnostics}
\label{app:robustness}

\begin{table}[t]
    \centering
    \small
    \caption{VarNet slice-level reconstruction NMSE under the matched source and target protocol. These values describe reconstruction error rather than risk-prediction RMSE.}
    \label{tab:varnet_operating_range}
    \setlength{\tabcolsep}{4pt}
    \begin{tabular}{lcccc}
        \toprule
        Split & Mean & Median & 5th pct. & 95th pct. \\
        \midrule
        Source & $0.00666$ & $0.00609$ & $0.00358$ & $0.01142$ \\
        M4Raw & $2.438$ & $2.418$ & $1.650$ & $3.401$ \\
        \bottomrule
    \end{tabular}
\end{table}

Median log-NMSE changes from $-5.101$ on source to $0.883$ on M4Raw. This confirms that VarNet is evaluated under a severe frozen-reconstructor stress test. Recomputing NMSE directly from cached $q_T$ and targets gives maximum log-NMSE discrepancies of $2.14\times10^{-8}$ on source and $2.22\times10^{-16}$ on M4Raw. The reversal is therefore not attributable to a stored-target mismatch.

For each calibration size $n\in\{4,8,16\}$, the low-shot diagnostic uses 1,000 patient-level random splits with seed 20260829. Calibration patients are selected without replacement from the 25 M4Raw patients. All scans from the selected patients are used to fit only the intercept and slope of the frozen trajectory score, and evaluation uses the remaining patients. Table~\ref{tab:varnet_repair} reports means and SDs across these repeated held-out splits.

\begin{table*}[t]
    \centering
    \footnotesize
    \caption{Matched VarNet low-shot affine repair. Values are mean $\pm$ SD across repeated patient-held-out splits. Difference is trajectory minus target-constant RMSE.}
    \label{tab:varnet_repair}
    \setlength{\tabcolsep}{5pt}
    \begin{tabular}{ccccc}
        \toprule
        Patients & Traj. RMSE & Constant RMSE & Difference & Test $r$ \\
        \midrule
        4 & $0.0924\pm0.0172$ & $0.2306\pm0.0270$ & $-0.1383\pm0.0282$ & $0.9348\pm0.0093$ \\
        8 & $0.0837\pm0.0100$ & $0.2205\pm0.0199$ & $-0.1368\pm0.0206$ & $0.9348\pm0.0148$ \\
        16 & $0.0795\pm0.0130$ & $0.2144\pm0.0272$ & $-0.1349\pm0.0285$ & $0.9327\pm0.0303$ \\
        \bottomrule
    \end{tabular}
\end{table*}

With four calibration patients, the fitted slope is $-0.2036\pm0.0366$ and its 2.5th to 97.5th split range is $[-0.2731,-0.1301]$. The trajectory-minus-constant RMSE range is $[-0.2042,-0.0846]$. These are empirical split ranges rather than confidence intervals. Repair fits only an offset and slope to the frozen trajectory coordinate.

\begin{table*}[t]
    \centering
    \footnotesize
    \caption{PromptMR+ scalar ResNet seed sensitivity. For each training seed, checkpoint selection uses source data only. Values are descriptive point estimates from the source holdout and zero-shot M4Raw evaluation.}
    \label{tab:resnet_seeds}
    \setlength{\tabcolsep}{4.2pt}
    \begin{tabular}{lcccccc}
        \toprule
        Seed & Source $r$ & Source $\rho_s$ & Source RMSE & Target $r$ & Target $\rho_s$ & Target RMSE \\
        \midrule
        Reported run & $0.187$ & $0.136$ & $0.492$ & $0.292$ & $0.313$ & $0.989$ \\
        20260816 & $0.261$ & $0.180$ & $0.457$ & $0.319$ & $0.369$ & $1.336$ \\
        20260817 & $0.477$ & $0.375$ & $1.076$ & $0.674$ & $0.715$ & $1.384$ \\
        \bottomrule
    \end{tabular}
\end{table*}

The endpoint ResNet is seed-sensitive in both ranking and calibration. The strongest additional target run remains below trajectory in correlation, while its absolute calibration is poor. The paper therefore treats this comparator as one endpoint diagnostic and bases the dynamics claim on initialization control, Hu-style conditional analysis, early-prefix prediction, and the broader uncertainty comparison.

\section{Discovery Model and Repeated Masks}
\label{app:discovery}

The earlier discovery-model experiment established the initial MRI risk predictor. It uses the separate descriptor defined in Sec.~\ref{app:features}. Table~\ref{tab:discovery_source} reports a fresh 100-volume evaluation. Trajectory improves calibrated prediction error over amplitude, while their Spearman difference interval includes zero. Adding endpoint information further reduces RMSE and MAE without a clearly resolved Spearman gain.

\begin{table*}[t]
    \centering
    \footnotesize
    \caption{Discovery-model source prediction and paired differences. Difference rows use the first method minus the second and include 95\% patient-bootstrap CIs. Inferential comparisons in the text use these paired intervals.}
    \label{tab:discovery_source}
    \setlength{\tabcolsep}{4pt}
    \begin{tabular}{lcccc}
        \toprule
        Predictor & $r$ & $\rho_s$ & RMSE & MAE \\
        \midrule
        Amplitude & $0.8930$ & $0.9521$ & $0.3873$ & $0.3278$ \\
        Trajectory & $0.9635$ & $0.9718$ & $0.2519$ & $0.1924$ \\
        Combined & $0.9777$ & $0.9787$ & $0.1760$ & $0.1390$ \\
        \midrule
        Difference & \multicolumn{2}{c}{$\Delta\rho_s$ and 95\% CI} & $\Delta$ RMSE and CI & $\Delta$ MAE and CI \\
        Trajectory $-$ amplitude & \multicolumn{2}{c}{$+0.0202$ $[-0.0049,0.0528]$} & $-0.1359$ $[-0.1987,-0.0794]$ & $-0.1360$ $[-0.1868,-0.0870]$ \\
        Combined $-$ trajectory & \multicolumn{2}{c}{$+0.0074$ $[-0.0031,0.0185]$} & $-0.0758$ $[-0.1174,-0.0356]$ & $-0.0538$ $[-0.0886,-0.0198]$ \\
        \bottomrule
    \end{tabular}
\end{table*}

For this corrected fresh-cohort evaluation, the predictor is trained on 32 volumes, conformal calibration uses a disjoint set of 99 fresh volumes, and evaluation uses 100 fresh test volumes. Feature definitions and ridge regularization are frozen before the calibration and test cohorts are evaluated.

At 50\% selective coverage, retaining the half of test volumes assigned the lowest predicted trajectory risk gives mean NMSE equal to $0.4281$ of the full-test-set mean and 90th-percentile NMSE equal to $0.3604$ of the full-test-set 90th percentile. Combined trajectory and endpoint selection gives ratios $0.4257$ and $0.3763$. Under one-sided split conformal calibration at nominal 90\% coverage, trajectory reaches empirical coverage $0.94$ and mean case-specific upper bound equal to $0.7346$ of the marginal bound. The combined model reaches coverage $0.92$ and bound ratio $0.6181$. These are finite-cohort results. The trajectory interval is conservative on this cohort.

The complete repeated-mask collection contains 199 volumes, eight masks per volume, and 1592 reconstructions. It supports the descriptive variance decomposition. Total log-risk variance is $0.7281374201$. Mean within-volume variance is $0.0045746952$, equal to $0.6283\%$ of the total. Approximate between-volume variance is $0.7236$. Mean maximum-to-minimum NMSE ratio over masks is $1.21277$. Inferential associations use the previously frozen 100-volume test subset, which contains 800 volume-mask reconstructions.

\begin{table*}[t]
    \centering
    \footnotesize
    \caption{Discovery-model repeated-mask Spearman associations on the frozen 100-volume test subset. Values are estimate $\pm$ source-volume-cluster bootstrap SD with 95\% percentile CI from 5,000 replicates. Between averages masks within volume. Within centers predictions and targets within volume. Partial correlation additionally removes realized mask density. Mask density uses the cluster-LOO mask-fraction predictor.}
    \label{tab:discovery_masks_full}
    \setlength{\tabcolsep}{4pt}
    \begin{tabular}{lcccc}
        \toprule
        Score & Pooled $\rho_s$ & Between $\rho_s$ & Within $\rho_s$ & Partial $\rho_s$ \\
        \midrule
        Amplitude & \shortstack{$0.9505\pm0.0128$\\$[0.9172,0.9662]$} & \shortstack{$0.9507\pm0.0139$\\$[0.9147,0.9681]$} & \shortstack{$0.6704\pm0.0347$\\$[0.5980,0.7354]$} & \shortstack{$0.4066\pm0.0491$\\$[0.3127,0.5046]$} \\
        Trajectory & \shortstack{$0.9660\pm0.0062$\\$[0.9480,0.9720]$} & \shortstack{$0.9763\pm0.0056$\\$[0.9604,0.9825]$} & \shortstack{$-0.1205\pm0.0405$\\$[-0.2004,-0.0419]$} & \shortstack{$0.1419\pm0.0412$\\$[0.0593,0.2215]$} \\
        Mask density & -- & -- & \shortstack{$0.6929\pm0.0168$\\$[0.6587,0.7247]$} & -- \\
        \bottomrule
    \end{tabular}
\end{table*}

The trajectory strongly orders physical volumes and provides little mask-specific ranking within a volume. Endpoint amplitude and realized density capture more of the fine mask variation. Under this repeated-mask protocol, the discovery-model trajectory score primarily reflects persistent case difficulty rather than fine mask-specific variation. This decomposition is specific to the repeated-mask protocol and the discovery-model descriptor. It is not an estimate of the between-case and within-case decomposition for the matched shared descriptor.

\section{Severe Shift, Group Structure, and Calibration}
\label{app:shift}

For both group-structure analyses, the pooled model regresses log-NMSE on the frozen source score. The patient+acquisition model additionally adjusts for patient and acquisition effects. The strict model further adjusts for endpoint amplitude and mask density.

The older PromptMR stress test changes source anatomy from fastMRI knee to M4Raw brain and includes multiple target acquisitions. Table~\ref{tab:orientation_full} reports adaptation and disjoint confirmation cohorts. Pooled and patient-and-acquisition adjusted slopes are negative in both. The strict confirmation interval reaches zero at its upper endpoint.

\begin{table}[t]
    \centering
    \footnotesize
    \caption{PromptMR severe-shift slopes with 95\% patient-bootstrap CIs. Strict models add endpoint amplitude and mask density to patient and acquisition effects.}
    \label{tab:orientation_full}
    \setlength{\tabcolsep}{3.2pt}
    \begin{tabular}{llcc}
        \toprule
        Cohort & Model & Slope & 95\% CI \\
        \midrule
        Adaptation & Pooled & $-0.916$ & $[-1.254,-0.577]$ \\
        & Patient+acquisition & $-0.546$ & $[-0.700,-0.399]$ \\
        & Strict & $-0.434$ & $[-0.615,-0.246]$ \\
        \midrule
        Confirmation & Pooled & $-1.383$ & $[-1.641,-1.127]$ \\
        & Patient+acquisition & $-0.323$ & $[-0.572,-0.122]$ \\
        & Strict & $-0.187$ & $[-0.438,0.0002]$ \\
        \bottomrule
    \end{tabular}
\end{table}

An older XPDNet group-structure analysis provides a complementary case. Its source slope is $1.083$ with CI $[0.944,1.226]$. M4Raw pooled slope is $1.560$ with CI $[1.484,1.642]$, Pearson $0.857$, and Spearman $0.841$. Patient-and-acquisition adjustment reduces the slope to $0.0935$ with CI $[0.0271,0.1622]$. The strict slope is $0.0641$ with CI $[-0.0005,0.1337]$. The positive pooled relation is therefore strongly associated with patient and acquisition structure. This finding motivates reporting pooled utility together with group-adjusted analysis.

Low-shot PromptMR calibration fits only the intercept and slope of the frozen source coordinate. Each calibration model is evaluated on the disjoint 25-patient confirmation cohort. Table~\ref{tab:calibration_full} reports all point estimates and paired intervals. The trajectory improves over a target constant for all cohort sizes. It improves over calibrated amplitude at 4 and 8 patients, while the 16-patient interval includes zero.

\begin{table*}[t]
    \centering
    \footnotesize
    \caption{Low-shot affine calibration on the severe PromptMR shift. Paired differences are trajectory minus comparator RMSE with 95\% patient-bootstrap CI.}
    \label{tab:calibration_full}
    \setlength{\tabcolsep}{3.6pt}
    \begin{tabular}{ccccccc}
        \toprule
        Patients & Constant & Amplitude & Trajectory & Traj. $-$ constant & Traj. $-$ amplitude & Reduction \\
        \midrule
        4 & $0.6921$ & $0.6810$ & $0.6464$ & $-0.0457$ $[-0.0618,-0.0301]$ & $-0.0346$ $[-0.0556,-0.0140]$ & $6.6\%$ \\
        8 & $0.6888$ & $0.6777$ & $0.6384$ & $-0.0504$ $[-0.0702,-0.0312]$ & $-0.0393$ $[-0.0774,-0.0014]$ & $7.3\%$ \\
        16 & $0.6883$ & $0.6778$ & $0.6452$ & $-0.0431$ $[-0.0577,-0.0290]$ & $-0.0325$ $[-0.0682,0.0034]$ & $6.3\%$ \\
        \bottomrule
    \end{tabular}
\end{table*}

\section{Controlled-Corruption Pilot and Scope}
\label{app:pilot}

The MRI risk method was preceded by a controlled study of intermediate states in an image restoration transformer. Corruption severity varied while image content and network parameters remained fixed. For undersampling, directions associated with approximately 80-line and 40-line conditions have cosine $0.95$--$0.97$. Clean-conditioned direction prediction reaches cosine $0.881762$ at down1/block0 and $0.877843$ at down1/block1.

The directional result was replicated over eight held-out images with three realizations per image. Across this replication, the predicted relation direction has cosine $0.5156\pm0.0893$ with the held-out relation direction, but only $0.000503\pm0.000583$ cosine with the negative reconstruction-loss gradient. A separate single-case leverage probe at $\beta=0.1$ confirms the causal distinction: a matched true-gradient step improves PSNR by $0.00855$ dB, whereas stepping opposite the predicted relation direction changes PSNR by only $0.000047$ dB.

The replicated intervention audit uses $\beta=0.01$. Across the eight-image, three-realization replication, finite-difference gradient descent improves PSNR by $0.003988$ dB on average and is positive in every tested condition, while gradient ascent is negative in every tested condition. Predicted relation reversal changes PSNR by approximately $1.09\times10^{-5}$ dB on average, is positive for only $4/8$ images, and passes the paired directional criterion in $5/24$ image-realization pairs. Performance relative to matched random controls is approximately at chance. Together with the single-case probe, the replicated experiment separates predictable representation motion from reconstruction-loss descent.

\begin{table}[t]
    \centering
    \small
    \caption{Controlled-corruption pilot. The first two rows summarize the eight-image, three-realization directional replication. The final two rows are from the separate single-case leverage probe at $\beta=0.1$.}
    \label{tab:pilot}
    \setlength{\tabcolsep}{3pt}
    \begin{tabular}{p{47mm}c}
        \toprule
        Measurement & Value \\
        \midrule
        Replicated relation-direction cosine & $0.5156\pm0.0893$ \\
        Cosine with negative loss gradient & $0.000503\pm0.000583$ \\
        Single-case true-gradient PSNR change & $+0.00855$ dB \\
        Single-case reverse predicted-direction PSNR change & $+0.000047$ dB \\
        \bottomrule
    \end{tabular}
\end{table}

The pilot supports an observational interpretation. Intermediate paths can encode degradation severity even when their tangent provides no useful loss-reduction direction. The MRI estimator reads the trajectory as evidence about risk and makes no intervention claim.

The matched study covers three reconstruction backbones, one source dataset, one target dataset, fixed nominal acceleration, and log-NMSE risk. It does not establish universal zero-shot transfer. PromptMR+, XPDNet, and VarNet show positive, attenuated, and reversed readouts under the same target protocol. Physics-side residuals can dominate under some shifts. Prospective deployment therefore requires signed target validation and, when labels are available, calibration at the patient level.
\end{document}

%% file: preamble.tex
\newcommand{\rms}{\operatorname{RMS}}
\newcommand{\nmse}{\operatorname{NMSE}}